\documentclass[conference]{IEEEtran}
\IEEEoverridecommandlockouts
\usepackage{cite}
\usepackage{amsmath,amssymb,amsfonts}
\usepackage{algorithmic}
\usepackage{graphicx}
\usepackage{stfloats}
\usepackage{textcomp}
\usepackage[bookmarks=false]{hyperref}
\hypersetup{
    colorlinks = true,
    citecolor  = blue,
    linkcolor  = blue,
    urlcolor   = blue,
}
\usepackage{colortbl}
\usepackage[table]{xcolor}
\usepackage{multirow}
\usepackage{booktabs}
\usepackage{cleveref}
\usepackage{subcaption}
\def\equationautorefname~#1\null{Eq.~(#1)\null}
\def\figureautorefname~#1\null{Fig.~#1\null}
\def\tableautorefname~#1\null{Tab.~#1\null}
\def\definitionautorefname~#1\null{Def.~#1\null}
\def\sectionautorefname~#1\null{Sect.~#1\null}
\def\subsectionautorefname~#1\null{Sect.~#1\null}
\def\subsubsectionautorefname~#1\null{Sect.~#1\null}

\newcommand{\ultra}{\textsf{\footnotesize{ULTRA8T}}}
\newcommand{\sandwich}{\textsf{\footnotesize{SANDWICH-RAM}}}
\newcommand{\ssram}{\textsf{\footnotesize{SSRAM}}}
\newcommand{\digtime}{\textsf{\footnotesize{DIGITAL\_CLK\_GEN}}}
\newcommand{\timectrl}{\textsf{\footnotesize{TIMING\_CTRL}}}
\newcommand{\sarray}{\textsf{\footnotesize{ARRAY\_128\_32}}}

\def\BibTeX{{\rm B\kern-.05em{\sc i\kern-.025em b}\kern-.08em
    T\kern-.1667em\lower.7ex\hbox{E}\kern-.125emX}}
\begin{document}

\title{\huge
    Transferable Parasitic Estimation via Graph Contrastive Learning and Label Rebalancing in AMS Circuits\\
}


\author{\IEEEauthorblockN{Shan Shen$^{1,2}$, Shenglu Hua$^{3}$, Jiajun Zou$^{1}$, Jiawei Liu$^3$, Jianwang Zhai$^3$, Chuan Shi$^3$, and Wenjian Yu$^2$}
\IEEEauthorblockA{$^{1}$\textit{Nanjing University of Science and Technology, Nanjing 210094, China} \\
$^{2}$\textit{Tsinghua University, Beijing 100084, China} \\
$^{3}$\textit{Beijing University of Posts and Telecommunications, Beijing 100876, China}}
\thanks{
This work is supported by the National Key R\&D Program of China (No. 2022YFB2901100), the National Natural Science Foundation of China (NSFC) (No. 62204141, 62404021), and the Beijing Natural Science Foundation (No. Z230002, 4244107, QY24216, QY24204, QY25329). 
S. Shen and S. Hua contributed equally to this work.
J. Zhai and W. Yu are the corresponding authors. 
}
\vspace{-14pt}
}

\maketitle

\begin{abstract}

Graph representation learning on Analog-Mixed Signal (AMS) circuits is crucial for various downstream tasks, e.g., parasitic estimation.
However, the scarcity of design data, the unbalanced distribution of labels, and the inherent diversity of circuit implementations pose significant challenges to learning robust and transferable circuit representations.
To address these limitations, we propose CircuitGCL, a novel graph contrastive learning framework that integrates representation scattering and label rebalancing to enhance transferability across heterogeneous circuit graphs.
CircuitGCL employs a self-supervised strategy to learn topology-invariant node embeddings through hyperspherical representation scattering, eliminating dependency on large-scale data. 
Simultaneously, balanced mean squared error (BMSE) and balanced softmax cross-entropy (BSCE) losses are introduced to mitigate label distribution disparities between circuits, enabling robust and transferable parasitic estimation.
Evaluated on parasitic capacitance estimation (edge-level task) and ground capacitance classification (node-level task) across TSMC 28nm AMS designs, CircuitGCL outperforms all state-of-the-art (SOTA) methods, with the $R^2$ improvement of $33.64\% \sim 44.20\%$ for edge regression and F1-score gain of $0.9\times \sim 2.1\times$ for node classification. Our code is available at https://github.com/ShenShan123/CircuitGCL.

\end{abstract}


\section{Introduction}


Modern Analog-Mixed Signal (AMS) circuits, which integrate analog blocks (e.g., amplifiers, oscillators) with digital subsystems (e.g., controllers, SRAM arrays), demand extensive manual iterations during design. Engineers must balance conflicting requirements—analog components require precise tuning of electrical parameters (gain, linearity), while digital blocks prioritize timing closure and power efficiency. Post-layout parasitic effects (e.g., unintended capacitive coupling) further compound this complexity, often necessitating time-consuming revisions across schematic design, layout optimization, and iterative transistor-level simulations. For instance, in high-speed SRAM designs, coupling capacitance between adjacent interconnects can degrade signal integrity, requiring weeks of manual adjustments to meet yield targets.

Recently, Deep Learning (DL) methods based on Neural Networks (NN) have offered transformative solutions to reduce the design complexity of AMS circuits. 
Among them, Graph Neural Networks (GNNs) natively model circuits as graphs, where nodes represent components (i.e., transistors, nets) and edges encode connectivity or coupling effects \cite{ParaGraph, shen2024deep, shen2025few-shot, yu2025deep}. 
By treating parasitics as learnable edge or node attributes, this enables pre-layout prediction of parasitic capacitance (a task conventionally deferred to post-layout verification), significantly reducing the need for iterative layout-simulation loops. 

However, high-quality AMS circuit data, including SPICE netlists, layout parasitics, and performance metrics, is often proprietary and expensive to generate. This \textit{scarcity} limits the adoption of large-scale GNNs (e.g., deep GNNs or graph transformers) in mixed-signal design flows. Consequently, supervised approaches struggle in this regime, resulting in overfitting and poor robustness when applied to unseen circuit topologies or advanced semiconductor technologies.

\begin{figure}[!t]
    \vspace{-6pt}
    \setlength{\abovecaptionskip}{-4pt}
    \setlength{\belowcaptionskip}{-4pt}
    \centering
    \includegraphics[width=0.9\linewidth]{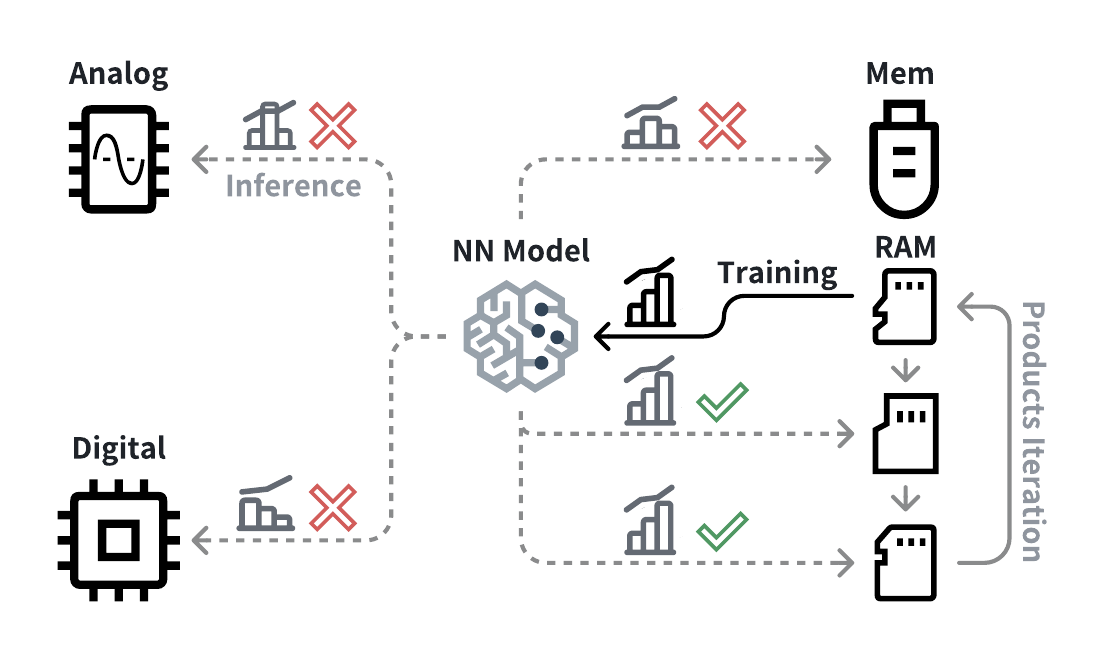}
    \caption{NN-based DL model has poor transferability due to circuit heterogeneity. }
    \label{fig:motive}
\end{figure}

Furthermore, AMS circuits exhibit inherent \textit{diversity}, spanning analog, digital, and mixed-signal domains, as illustrated in \autoref{fig:motive}. 
Each type of circuit is governed by distinct design principles and performance requirements. Memory circuits, which integrate analog and digital subsystems, exacerbate this variability. 
Traditional NNs trained on specific circuit types often fail to generalize to others due to differences in topology, operating regimes, and optimization objectives. 
While they show promise for predicting similar designs or performing iterative tasks within a product family, the lack of cross-domain transferability often leads to costly retraining or dataset regeneration, thereby limiting the extensive usage of DL-driven Electronic Design Automation (EDA) tools. 

The combination of scarcity and diversity in circuit data further leads to a label \textit{imbalance} problem, which fundamentally constrains the transferability of GNNs. This imbalance is inherently prevalent in AMS circuit datasets, where label distributions are often skewed with long-tailed patterns rather than uniformly distributed across categories. For instance, parasitic capacitors with larger capacitance, which result in severe timing violations and signal integrity, are significantly underrepresented, as reported in recent studies \cite{ParaGraph, shen2024deep}. Such imbalance significantly undermines both the generalizability and fairness of Graph Neural Networks (GNNs), presenting a critical challenge that demands dedicated efforts
to enhance data-driven transferability.

\textit{Can state-of-the-art transfer learning techniques address data scarcity and diversity in the EDA domain?}  
In this work, we answer this question affirmatively by proposing \textbf{CircuitGCL}, a framework that integrates a Representation Scattering Mechanism (RSM) into Graph Contrastive Learning (GCL) and employs label rebalancing techniques.  
Specifically, we focus on parasitic capacitance estimation at the pre-layout stage.
By modeling circuit nets as nodes and coupling effects as edges, we evaluate the proposed method on two downstream tasks: (i) \textit{edge regression} to estimate coupling capacitance values and (ii) \textit{node classification} to categorize the ground capacitance of each net into discrete ranges (small/medium/large).  

Our key contributions, summarized in \autoref{fig:workflow}, are as follows:  

\begin{itemize}  
    \item We adapt the Representation Scattering Mechanism (RSM) for GCL and demonstrate that it generates transferable representations for various circuit graphs. These representations are directly applicable to other unseen AMS designs without any task-specific fine-tuning.  
    \item We address data imbalance in circuit datasets through label rebalancing, enhancing model transferability across domains. For regression tasks, we adopt balanced Mean Squared Error (MSE), while balanced softmax Cross-Entropy (BSCE)  is applied to classification tasks.  
    \item We deem the above two contributions make CircuitGCL extend directly to resistance/inductance prediction, crosstalk analysis, IR drop estimation, and cross-technology transfer.
\end{itemize}

\section{Preliminary}

We first introduce the basic task-related background of parasitic estimation and some preliminaries about GCL and imbalanced classification and regression.

\subsection{Parasitic Capacitance Estimation}
The design of AMS circuits typically requires extensive manual intervention, relying heavily on iterative topology selection and component sizing. 
In traditional workflows, IC engineers optimize circuits through pre-layout simulations and verify designs using post-layout simulations. However, as technology scales to advanced nodes, reduced feature sizes, tighter spacing, and lower supply voltages collectively amplify parasitic effects. These effects introduce significant discrepancies between pre-layout and post-layout simulation results, with parasitic capacitance emerging as a critical factor that can no longer be neglected during early design stages \cite{yu2014advanced,yu2009variational,yu2004preconditioned}.  
Parasitic capacitance arises from unintended capacitive coupling between conductive structures, electric field penetration through dielectrics, and non-ideal charge accumulation at interfaces. It is categorized into two types: (1) ground capacitance (between interconnects and the substrate) and (2) coupling capacitance (between adjacent interconnects). These parasitics degrade circuit performance by increasing propagation delays, raising power consumption, and compromising signal integrity.  

To address these challenges, graph neural networks have been adopted for parasitic capacitance prediction. For instance, ParaGraph \cite{ParaGraph} converts circuit schematics into graphs and employs message-passing neural network (MPNN) layers to predict net capacitance and layout parameters. The framework uses three ensemble models, selecting the best-performing output to mitigate label imbalance in lumped capacitance predictions. Similarly, Shen et al. \cite{shen2024deep} developed DLPL-Cap, a deep learning model combining a GNN router with five expert regressors to handle imbalanced data distributions in SRAM circuits during pre-layout parasitic estimation. However, both works treat coupling capacitance as a component of lumped capacitance, limiting their granularity.  

Recent advancements, such as CircuitGPS \cite{shen2025few-shot}, propose few-shot learning for parasitic prediction using small-hop subgraph sampling, a low-cost positional encoding (double-anchor shortest path distance, DSPD), and pre-training strategies. While DSPD reduces computational complexity compared to traditional positional encodings, its time and storage costs scale poorly with graph size, necessitating restrictive 1-hop subgraph sampling. This limitation inspired our use of GCL to generate initial node embeddings. Additionally, CircuitGPS manually constructs negative edge samples without accounting for imbalanced label distributions — a gap our method explicitly tackles.  

\subsection{Graph Conversion of AMS Circuits}
The schematic netlist of an AMS circuit is modeled as a heterogeneous graph $\mathcal{G=(V, E)}$, where $\mathcal{V}=\{v_1, v_2, \dots, v_N\}$ represents the set of nodes with attribute matrix $\mathbf{X} \in \mathbb{R}^{N \times D}$, and $\mathcal{E \subseteq V \times V}$ denotes the set of edges. The adjacency matrix $\mathbf{A} \in \{0, 1\}^{N \times N}$ is defined such that $\mathbf{A}_{ij} = 1$ if an edge $(v_i, v_j) \in \mathcal{E}$ exists, and $\mathbf{A}_{ij} = 0$ otherwise. The degree matrix $\mathbf{D} \in \mathbb{R}^{N \times N}$ is diagonal, with entries $d_i = \sum_{j \in \mathcal{V}} \mathbf{A}_{ij}$ for each node $v_i$.  

Following the conversion from \cite{shen2025few-shot}, nodes in $\mathcal{V}$ are categorized into three types: nets, transistor devices, and pins (device terminals). Edges in $\mathcal{E}$ encode the schematic topology as either device-to-pin or net-to-pin connections. Coupling capacitance, which constitutes the prediction target, is excluded from $\mathcal{G}$ and modeled as candidate edges. These include three subtypes: pin-to-net, pin-to-pin, and net-to-net coupling. To simplify GNN models, heterogeneous AMS graphs are converted into homogeneous graphs by assigning node type attributes $\mathbf{X} \in \{0, 1, 2\}^{N \times 1}$, where each entry corresponds to a node type.  

To address the specific requirements of downstream regression tasks, particularly coupling capacitance prediction, the model incorporates an enhanced feature matrix $\mathbf{X}_C \in \mathbb{R}^{N \times d_C}$ that captures detailed design parameters and connectivity statistics. For net nodes, this matrix encodes comprehensive connectivity information, including the count and geometric properties of connected components such as transistors, capacitors, and resistors, along with their dimensional characteristics like width and length. Device nodes are characterized by their intrinsic parameters, including multiplier values, geometric dimensions, and device type codes. Pin nodes are distinguished by their functional roles within device instances, such as gate, drain, source, or bulk terminals in MOS devices. Notably, our method does not utilize edge attributes. 

\subsection{Graph Contrastive Learning}
The objective of graph contrastive learning is to train an encoder $f(\cdot)$ in a self-supervised manner. The learned encoder generates node representations $\boldsymbol{H} = f(\mathbf{X}, \mathbf{A})$, where $\mathbf{H} \in \mathbb{R}^{N \times k}$ captures both topological relationships and dense semantic patterns. These representations are decoupled from specific topological biases and can be generalized to diverse downstream tasks.

\begin{figure*}[!t]
\vspace{-4pt}
    \setlength{\abovecaptionskip}{0pt}
    \setlength{\belowcaptionskip}{-4pt}
    \centering
    \includegraphics[width=0.86\linewidth]{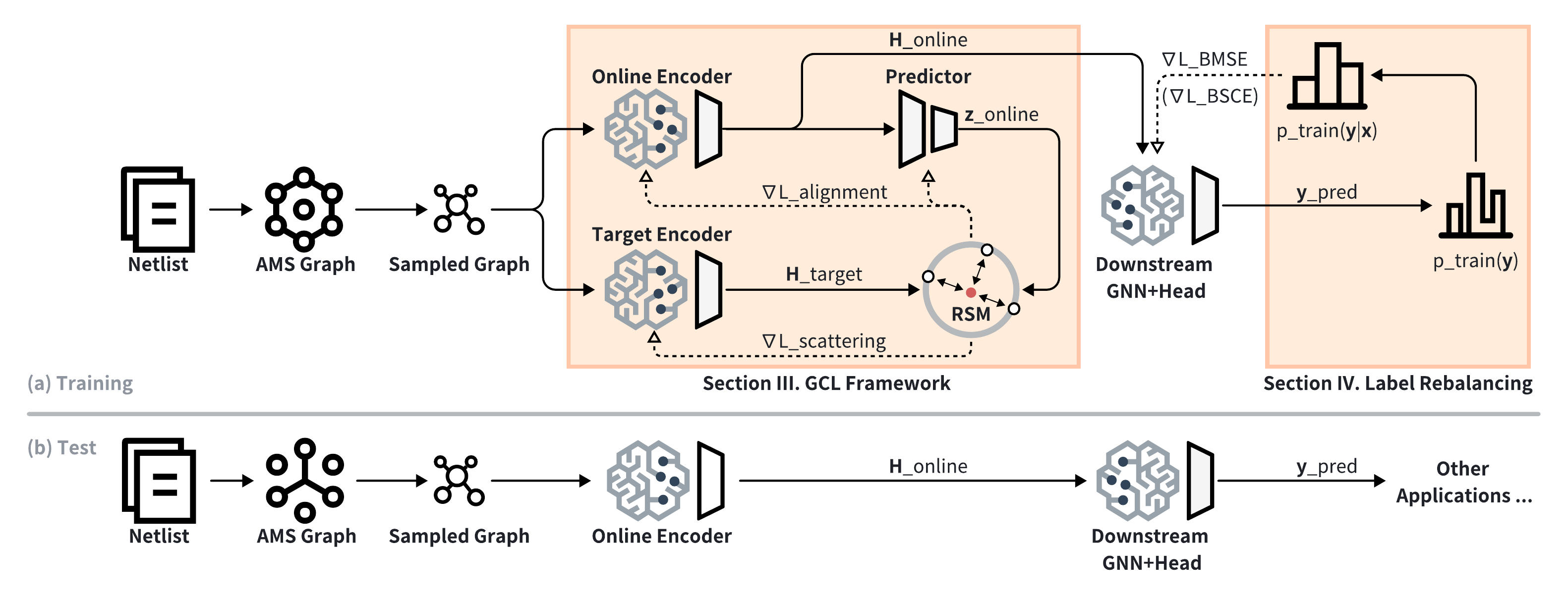}
    \caption{Workflow of CircuitGCL. (a) During training, a target encoder applies a Representation Scattering Mechanism (RSM) to generate scattered embeddings ($\mathbf{H}_{\text{target}}$), while an online encoder produces embeddings ($\mathbf{H}_{\text{online}}$) that are passed to a downstream GNN. To improve transferability, a label rebalancing module adjusts the final loss based on the training label distribution, $p_{\text{train}}(\boldsymbol{y})$. (b) During testing, only the trained online encoder and downstream GNN are utilized to generate predictions ($\boldsymbol{y}_{\text{pred}}$).}
    \label{fig:workflow}
\end{figure*}

He et al.~\cite{he2024exploitation} found that mainstream GCL frameworks \cite{zhu2020deep, zhu2021graph, xia2022progcl, zheng2022rethinking, velickovic2019deep, hassani2020contrastive, thakoor2021bootstrapped,lee2022augmentation, sun2024rethinking} all inherently perform representation scattering, which plays a crucial role in their success. Here, we follow the definition of Representation Scattering in their work:
\newtheorem{definition}{Definition}
\begin{definition}[Representation Scattering]
\label{def:rs}
In a $D$-dimensional embedding space $\mathbb{R}^D$ with $N$ node embeddings represented as $\mathbf{X} \in \mathbb{R}^{N \times D}$, let $\mathbb{S}^k$ ($1 \leq k \leq D$) denotes a subspace of $\mathbb{R}^D$ and $\boldsymbol{c}$ denotes a scatter center. Representation scattering enforces two constraints:  
(i) center-away constraint, where node embeddings are maximally separated from $\boldsymbol{c}$;  
(ii) uniformity constraint, where node embeddings are uniformly distributed across $\mathbb{S}^k$.  
\end{definition}

According to \autoref{def:rs}, representation scattering requires defining a scatter center $\boldsymbol{c}$ within the subspace $\mathbb{S}^k$ and enforcing both center-away and uniformity constraints. 
Such a mechanism constructs a common space for nodes from different circuit graphs, and also normalizes those representations.

\subsection{Imbalanced Classification and Regression}
In machine learning, label imbalance poses significant challenges for deep recognition models, motivating numerous techniques to address data imbalance \cite{ren2020balanced, menon2020long, hong2021disentangling, ren2022balanced, kang2019decoupling, zhou2020bbn, wang2020long, steininger2021density, byrd2019effect, xu2021understanding}.  
Most prior works focus on imbalanced classification (also termed long-tailed recognition \cite{liu2019large}), with solutions broadly categorized into:  
(i) data-based methods, such as oversampling minority classes or undersampling majority classes \cite{chawla2002smote};  
(ii) model-based methods, including loss reweighting or adjusted objective functions to mitigate class imbalance \cite{cao2019learning, ren2020balanced}.  

In contrast, many EDA tasks involve regression with continuous, unbounded target values, where label imbalance is inherently more challenging. 
For example, Shen et al. \cite{shen2024deep} reported that the distribution of ground parasitic capacitance spans from 0.01 fF to 100 pF, where over 10\textsuperscript{6} samples fall into the second bin [10\textsuperscript{-1}, 1].
While imbalanced classification is well-studied, imbalanced regression remains under-explored. Existing approaches often adapt the synthetic
 minority over-sampling technique (SMOTE) to regression scenarios \cite{torgo2013smote, branco2017smogn} or employ loss reweighting strategies \cite{yang2021delving, steininger2021density}. Reweighting assigns higher loss weights to rare samples and lower weights to frequent ones. However, recent studies \cite{byrd2019effect, xu2021understanding, ren2022balanced} demonstrate limited effectiveness of reweighting in both classification and regression tasks.  

Consider input $\boldsymbol{x} \in \mathbf{X}$ and label $\boldsymbol{y} \in \mathbf{Y} = \mathbb{R}^d$. 
We focus on univariate regression ($d=1$), where training and test data originate from distinct AMS designs. 
This implies that the training set follows a skewed joint distribution $p_{\text{train}}(\boldsymbol{x}, \boldsymbol{y})$, while the test set adheres to a near-uniform or lightly skewed distribution $p_{\text{bal}}(\boldsymbol{x}, \boldsymbol{y})$ \cite{yang2021delving, ren2022balanced}. 
Crucially, the label-conditional distribution $p(\boldsymbol{x} | \boldsymbol{y})$ is assumed invariant between training and testing. Under the assumption of a balanced test set, the goal of imbalanced regression shifts from estimating $p_{\text{train}}(\boldsymbol{y} | \boldsymbol{x})$ to learning $p_{\text{bal}}(\boldsymbol{y} | \boldsymbol{x})$, ensuring generalizability to unseen circuits. 
This equivalence aligns with the theoretical insight that balanced metrics on arbitrary test sets mirror overall metrics on hypothetical balanced sets \cite{brodersen2010balanced}.  

The mean squared error (MSE) loss is the most widely used objective function in regression tasks. For a predicted value $\boldsymbol{y}_{\text{pred}}$ and target $\boldsymbol{y}$, the MSE loss is defined as:
\begin{equation}
\text{MSE}(\boldsymbol{y}, \boldsymbol{y}_{\text{pred}}) = \left\| \boldsymbol{y} - \boldsymbol{y}_{\text{pred}} \right\|^2_2, \label{eq:mse}
\end{equation}
where $\left\| \cdot \right\|_2$ denotes the $\ell_2$-norm. 
From a probabilistic perspective, the prediction $\boldsymbol{y}_{\text{pred}}$ can be interpreted as the mean of a Gaussian distribution modeling the prediction noise \cite{mccullagh2019generalized}:
\begin{align}\label{eq:gaussian}
p(\boldsymbol{y} | \boldsymbol{x}; \boldsymbol{\theta}) = \mathcal{N}(\boldsymbol{y}; \boldsymbol{y}_{\text{pred}}, \sigma_{\text{noise}}^2 \mathbf{I}),
\end{align}
where $\boldsymbol{\theta}$ represents the regressor's parameters, and $\sigma_{\text{noise}}^2 \mathbf{I}$ is the covariance matrix of the independent and identically distributed (i.i.d.) error term $\epsilon \sim \mathcal{N}(0, \sigma_{\text{noise}}^2 \mathbf{I})$. The MSE loss is equivalent to the Negative Log-Likelihood (NLL) of this Gaussian distribution \cite{nix1994estimating}, implying that MSE-trained regressors inherently model $p_{\text{train}}(\boldsymbol{y} | \boldsymbol{x})$.
In order to enhance performance on testsets, we aim to estimate $p_{\text{bal}}(\boldsymbol{y} | \boldsymbol{x})$ instead of $p_{\text{train}}(\boldsymbol{y} | \boldsymbol{x})$. 
Bayes' theorem is applied:
\begin{align}\label{eq:change_variable}
\frac{p_{\text{train}}(\boldsymbol{y} | \boldsymbol{x})}{p_{\text{bal}}(\boldsymbol{y} | \boldsymbol{x})} \propto 
\frac{p(\boldsymbol{x} | \boldsymbol{y}) \cdot p_{\text{train}}(\boldsymbol{y})}{p(\boldsymbol{x} | \boldsymbol{y}) \cdot p_{\text{bal}}(\boldsymbol{y})} 
= \frac{p_{\text{train}}(\boldsymbol{y})}{p_{\text{bal}}(\boldsymbol{y})}.
\end{align}
This proportionality reveals that the discrepancy between $p_{\text{train}}(\boldsymbol{y} | \boldsymbol{x})$ and $p_{\text{bal}}(\boldsymbol{y} | \boldsymbol{x})$ depends on the ratio of their label distributions. Since $p_{\text{train}}(\boldsymbol{y})$ is lower for rare labels, MSE-trained regressors systematically underestimate underrepresented targets in the training set.

\section{The Proposed CircuitGCL Framework}
Overall workflow of CircuitGCL is depicted by Fig. \ref{fig:workflow}. 
The proposed method contains four steps: (i) AMS netlist conversion, which is the same as work \cite{shen2025few-shot}; (ii) subgraph sampling; (iii) representation scattering in GCL; and (iv) label rebalancing through balanced MSE and balanced softmax cross-entropy (BSCE). 
In this section, we try to tackle the aforementioned scarcity and diversity of AMS circuits by introducing contrastive learning in the parasitic estimation field.

\subsection{Representation Scattering Mechanism in GCL} 
To directly predict parasitic parameters from circuit topology, the absence of layout and detailed circuit information severely constrains the feature space. We employ contrastive learning to derive meaningful initial feature representations, while its self-supervised nature enables effective cross-domain transferability. Since current mainstream graph contrastive learning methods inherently employ representation scattering mechanisms (RSM), we adopt Scattering Graph Representation Learning (SGRL) \cite{he2024exploitation} as our GCL foundation, which embeds node representations within a hypersphere while dispersing them from a central mean point (as depicted in \autoref{fig:workflow}), providing a bias-free method that preserves circuit structure and attributes.

RSM operates by defining a subspace $\mathbb{S}^k$ and a scatter center $\boldsymbol{c}$. To project representations from the original space $\mathbb{R}^D$ into $\mathbb{S}^k$, a transformation function $\text{Trans}(\cdot)$ applies $\ell_2$-normalization to each row vector $\boldsymbol{h}_i$ in the target representation matrix $\mathbf{H}_{\text{target}}$:
\begin{equation}
\widetilde{\boldsymbol{h}}_i = \frac{\boldsymbol{h}_i}{\max\left(\left\| \boldsymbol{h}_i \right\|_2, \varepsilon\right)}, \quad 
\mathbb{S}^k = \left\{ \widetilde{\boldsymbol{h}}_i : \left\| \widetilde{\boldsymbol{h}}_i \right\|_2 = 1 \right\},
\label{eq:h_tilde}
\end{equation}
where $\boldsymbol{h}_i$ is the representation of node $v_i \in \mathcal{V}$ generated by the target encoder, $\left\| \widetilde{\boldsymbol{h}}_i \right\|_2 = \left( \sum_{j=1}^k \widetilde{h}_{ij}^2 \right)^{1/2}$ denotes the $\ell_2$-norm, and $\varepsilon$ is a small constant (e.g., $10^{-8}$) to prevent division by zero. As per \autoref{eq:h_tilde}, all node representations are constrained to the hypersphere $\mathbb{S}^k$, ensuring stable training by preventing uncontrolled scattering in the embedding space.  

Next, we define the scattered center $\boldsymbol{c}$ and introduce a representation scattering loss $\mathcal{L}_{\text{scattering}}$ to push node representations away from $\boldsymbol{c}$ in $\mathbb{S}^k$:
\begin{equation}
\mathcal{L}_{\text{scattering}} = -\frac{1}{N} \sum_{i=1}^{N} \left\| \widetilde{\boldsymbol{h}}_i - \boldsymbol{c} \right\|_2^2, \quad \boldsymbol{c} = \frac{1}{N}\sum_{i=1}^{N} \widetilde{\boldsymbol{h}}_i,
\label{eq:l_scatter}
\end{equation}
where $\boldsymbol{c}$ represents the mean of all normalized node embeddings. By minimizing $\mathcal{L}_{\text{scattering}}$, RSM enforces global uniformity of representations across the dataset without enforcing strict local uniformity. 

\begin{figure}[!tb]
\setlength{\belowcaptionskip}{-2pt}
\centering
\begin{subfigure}[b]{0.35\linewidth}
    \includegraphics[width=\textwidth]{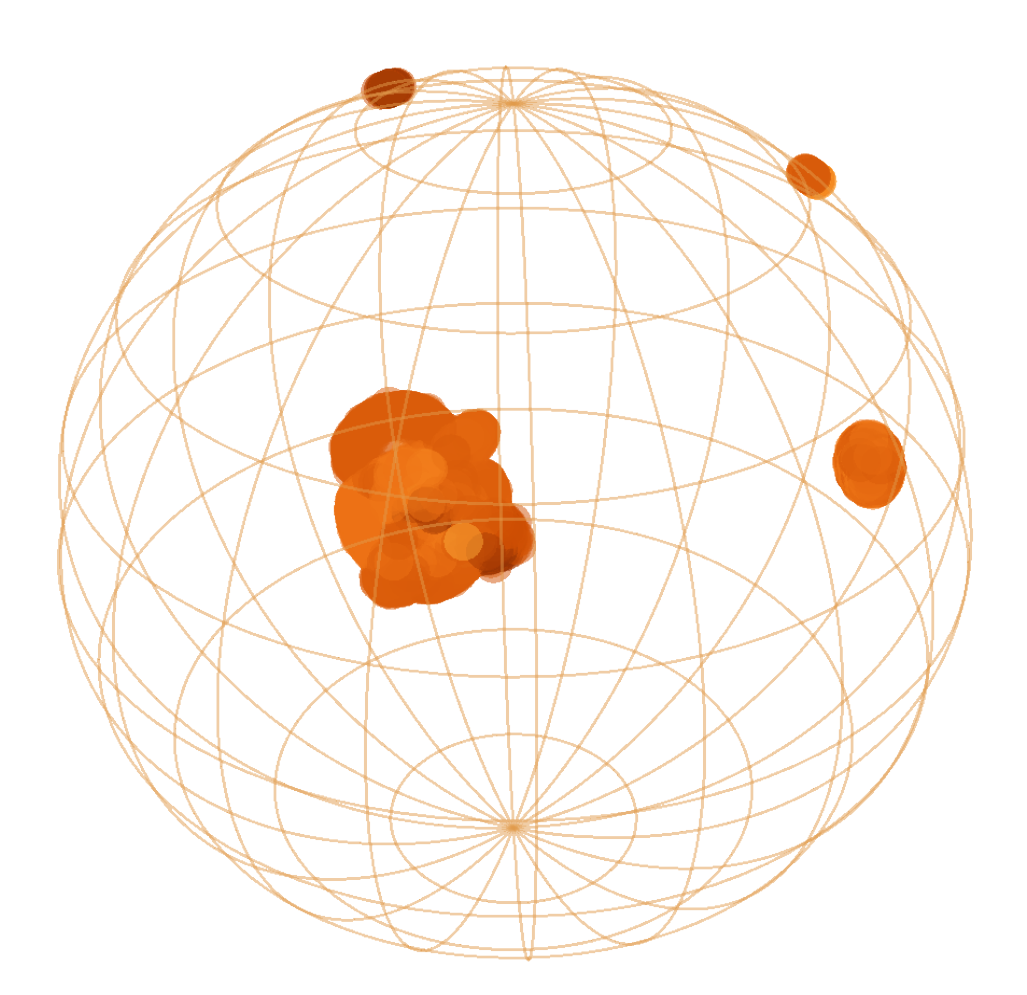}
    \caption{GNN without GCL.}
    \label{fig:tsne_sage}
\end{subfigure}
\quad
\begin{subfigure}[b]{0.35\linewidth}
    \includegraphics[width=\textwidth]{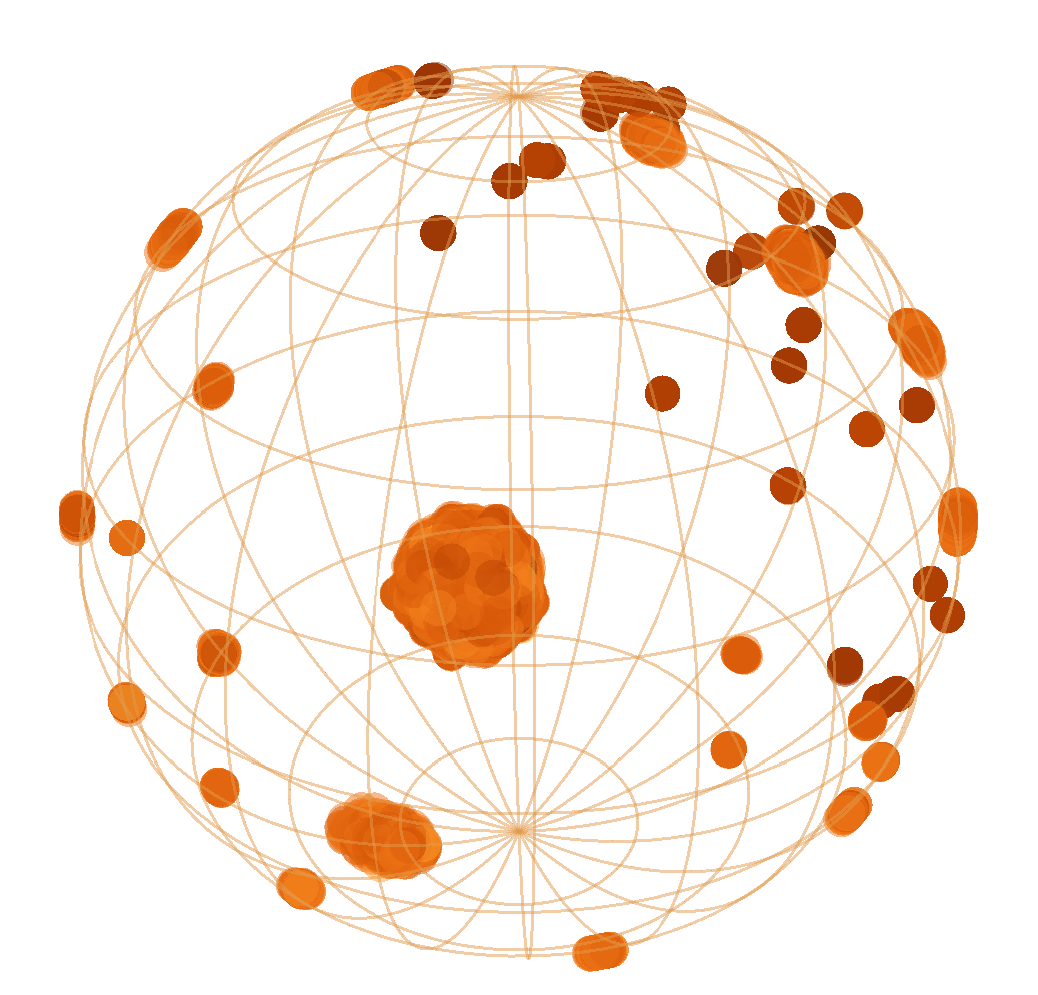}
    \caption{RSM of CircuitGCL}
    \label{fig:tsne_cirgcl}
\end{subfigure}
\caption{t-SNE visualizations of node embeddings: comparisons between models with and without the GCL framework. Darker indicates larger parasitic capacitance.}
\label{fig:tsne}
\end{figure}

\autoref{fig:tsne} visualizes the t-SNE embeddings of node representations learned by different contrastive learning frameworks.  
\autoref{fig:tsne_sage} depicts embeddings from a GNN trained solely on downstream tasks without pre-training. These embeddings lack structural and topological awareness, resulting in poor expressiveness that severely limits the transferability of conventional GNNs.  
In contrast, CircuitGCL’s RSM enforces uniform distribution across the subspace (\autoref{fig:tsne_cirgcl}), yielding highly expressive embeddings. Notably, RSM naturally clusters nodes with similar labels (lighter hues denote smaller parasitic capacitance values) while pushing nodes with dissimilar labels apart.  
This mechanism is particularly advantageous for circuit graph representation learning. While semantically distinct nodes (e.g., analog vs. digital components) are scattered across the hypersphere to maximize separation, functionally similar nodes (e.g., repeated inverter cells) naturally cluster in local regions. Consequently, RSM unifies embedding subspaces across diverse AMS designs, enabling robust transferability between circuits with varying topologies and label distributions.  

\subsection{Online Encoder}  
After generating the scattered representations $\mathbf{H}_{\text{target}} = f_\phi(\mathbf{A}, \mathbf{X})$ using the target encoder, the online encoder $f_\theta(\cdot)$ produces intermediate representations $\mathbf{H}_{\text{online}}$. These are passed through a predictor $q_\theta(\cdot)$ to obtain predicted representations $\boldsymbol{z}_{\text{online}} = q_\theta(\mathbf{H}_{\text{online}})$. The objective is to align $\boldsymbol{z}_{\text{online}}$ with $\mathbf{H}_{\text{target}}$, enhancing the model’s ability to capture semantically meaningful circuit patterns. The alignment loss $\mathcal{L}_{\text{alignment}}$ is defined as:  
\begin{equation}
\mathcal{L}_{\text{alignment}} = -\frac{1}{N} \sum_{i=1}^{N} \frac{\boldsymbol{z}_i^T \boldsymbol{h}_i}{\left\| \boldsymbol{z}_i \right\|_2 \left\| \boldsymbol{h}_i \right\|_2},  
\label{eq:l_align}  
\end{equation}  
where $\boldsymbol{z}_{\text{online}}$ and $\mathbf{H}_{\text{target}}$ denote predicted and target embeddings, respectively. During training, only the online encoder’s parameters $\boldsymbol{\theta}$ are updated via gradient descent, while the target encoder’s parameters $\boldsymbol{\phi}$ remain fixed.  

Unlike direct alignment of constrained and scattered representations, the predictor $q_\theta(\cdot)$ acts as an adaptive buffer, enabling the online encoder to learn stable, topology-aware embeddings.  

To ensure the target encoder incorporates topological semantics into the scattering process, rather than optimizing solely for uniformity, we update $\boldsymbol{\phi}$ using an exponential moving average (EMA) of $\boldsymbol{\theta}$ after each epoch:  
\begin{equation}
\boldsymbol{\phi} \leftarrow \tau \boldsymbol{\phi} + (1 - \tau) \boldsymbol{\theta},  
\label{eq:ema}  
\end{equation}  
where $\tau \in [0, 1]$ is a decay rate (typically $\tau \geq 0.99$). This gradual update prevents adversarial collapse between the encoders and stabilizes training.  

\begin{table}[tb]
\centering
\caption{Resource Usage of DSPD Calculation and GCL Training.} \label{tab:pe_vs_sgrl}
\setlength\tabcolsep{2pt}
\resizebox{\linewidth}{!}{
    \begin{tabular}{l|lll|lll|lll}
    \toprule 
    Dataset & \multicolumn{3}{c|}{\ssram} & \multicolumn{3}{c|}{\ultra} & \multicolumn{3}{c}{\sandwich}  \\ 
    \rowcolor{gray!40}
    Resource & Time & Mem. & Disk &Time & Mem. & Disk & Time & Mem. & Disk \\ \midrule
    DSPD & 20.4m & 0.2GB & 68.6MB & 907.9m & 3.1GB & 2.6GB & 1115m & 9.6GB & 2.6GB \\
    GCL & 4.4m & 3.8GB & 5.4MB & 109.7m & 4.6GB & 5.4MB & 94.7m & 3.0GB & 5.4MB \\ \bottomrule
    \end{tabular} 
}
\parbox{\linewidth}{
    \vspace{2pt}
    \footnotesize \textit{Note}: DSPD uses CPU memory, and GCL's pre-training uses GPU video memory. Unit `m' stands for minute.
}
\end{table}

\subsection{Comparison with DSPD}  
In CircuitGPS \cite{shen2025few-shot}, double-anchor shortest path distance (DSPD) serves as a critical positional encoding (PE) for initial node embeddings during pre-training and fine-tuning. DSPD computes the relative shortest path distances between nodes in subgraphs using the resources of CPUs, but its computational and storage costs scale quadratically with subgraph size. As shown in \autoref{tab:pe_vs_sgrl}, DSPD becomes prohibitively expensive for large circuits, restricting 1-hop subgraph sampling in CircuitGPS. 
By contrast, CircuitGCL pre-trains the encoders to generate the initial embeddings to replace the DSPD calculation with high parallelism and good model scalability.


\section{Label Rebalancing}  

\begin{figure*}[t]
\setlength{\belowcaptionskip}{0pt}
\centering
\begin{subfigure}[b]{0.16\textwidth} 
    \setlength{\abovecaptionskip}{-2pt}
    \includegraphics[width=\textwidth]{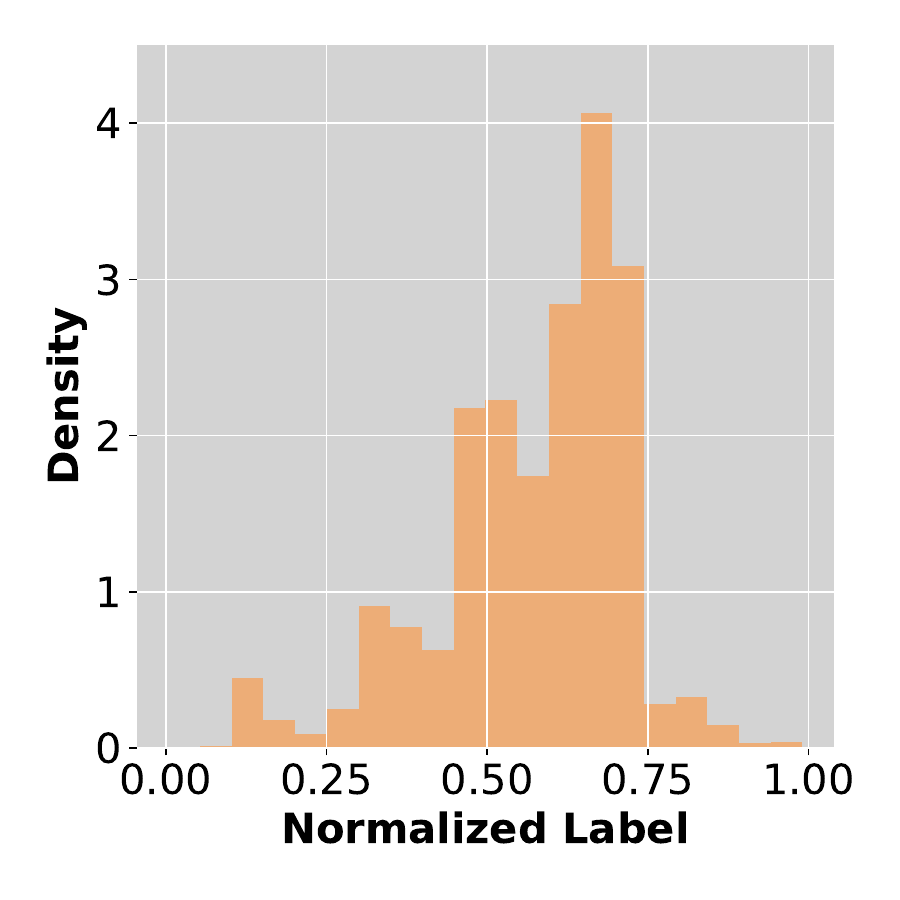}
    \caption{\textsf{\scriptsize{SSRAM}}}
    \label{fig:distr_a}
\end{subfigure}
\hfill 
\begin{subfigure}[b]{0.16\textwidth}
    \setlength{\abovecaptionskip}{-2pt}
    \includegraphics[width=\textwidth]{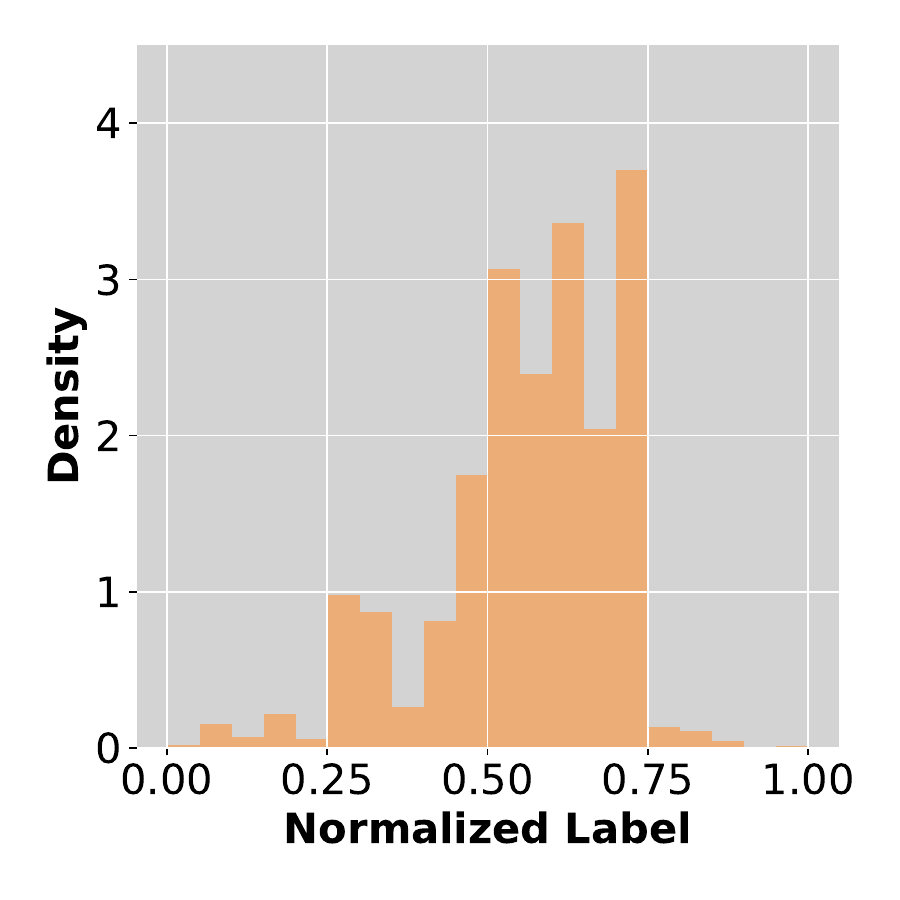}
    \caption{\textsf{\scriptsize{DIGITAL\_CLK\_GEN}}}
    \label{fig:distr_b}
\end{subfigure}
\hfill
\begin{subfigure}[b]{0.16\textwidth}
    \setlength{\abovecaptionskip}{-2pt}
    \includegraphics[width=\textwidth]{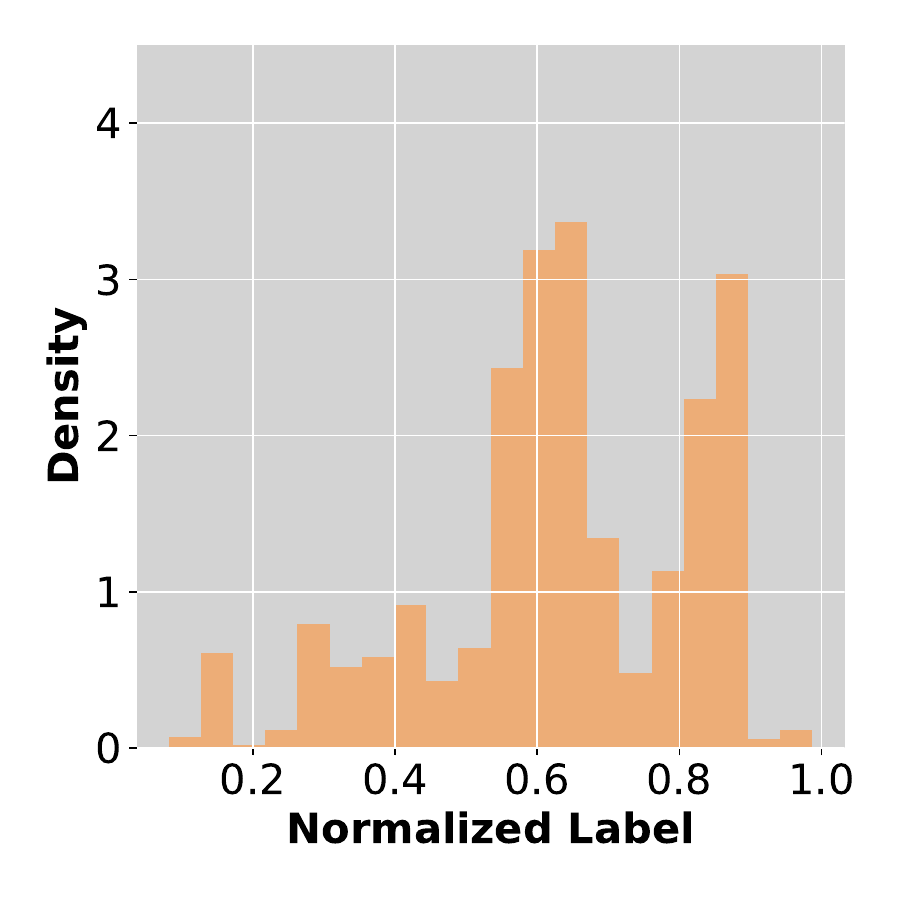}
    \caption{\scriptsize{\textsf{\scriptsize{TIMING\_CTRL}}}}
    \label{fig:1c}
\end{subfigure}
\hfill
\begin{subfigure}[b]{0.16\textwidth}
    \setlength{\abovecaptionskip}{-2pt}
    \includegraphics[width=\textwidth]{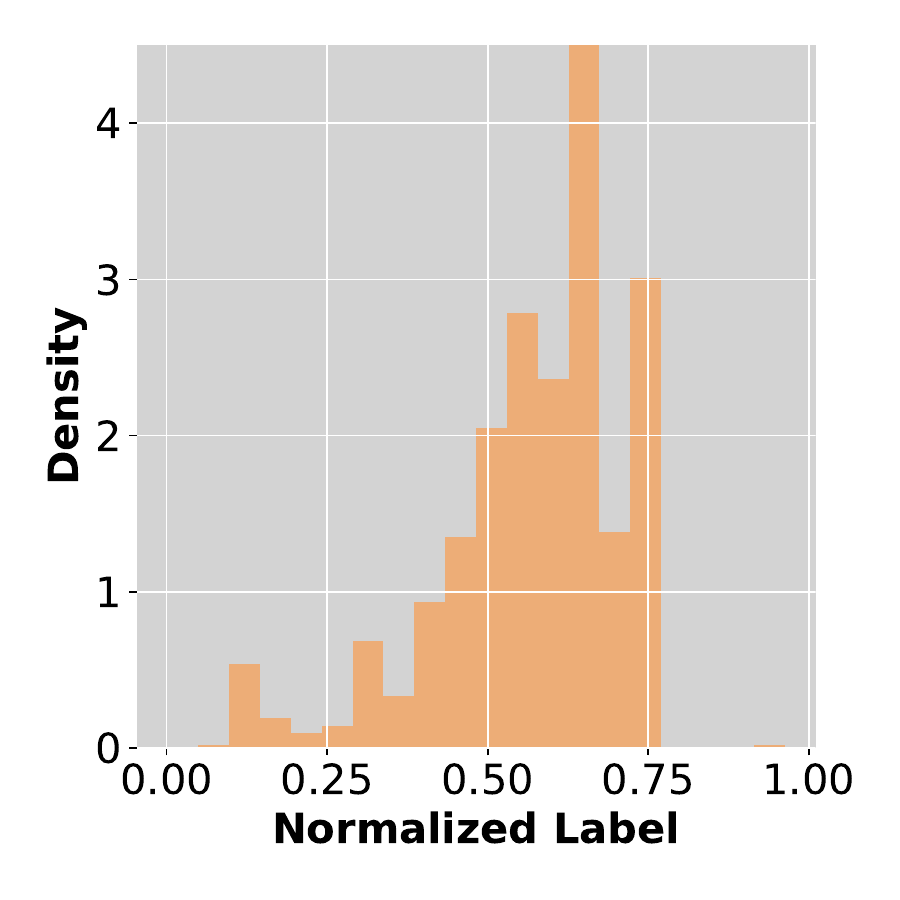}
    \caption{\textsf{\scriptsize{ARRAY\_128\_32}}}
    \label{fig:1d}
\end{subfigure}
\hfill
\begin{subfigure}[b]{0.16 \textwidth}
    \setlength{\abovecaptionskip}{-2pt}
    \includegraphics[width=\textwidth]{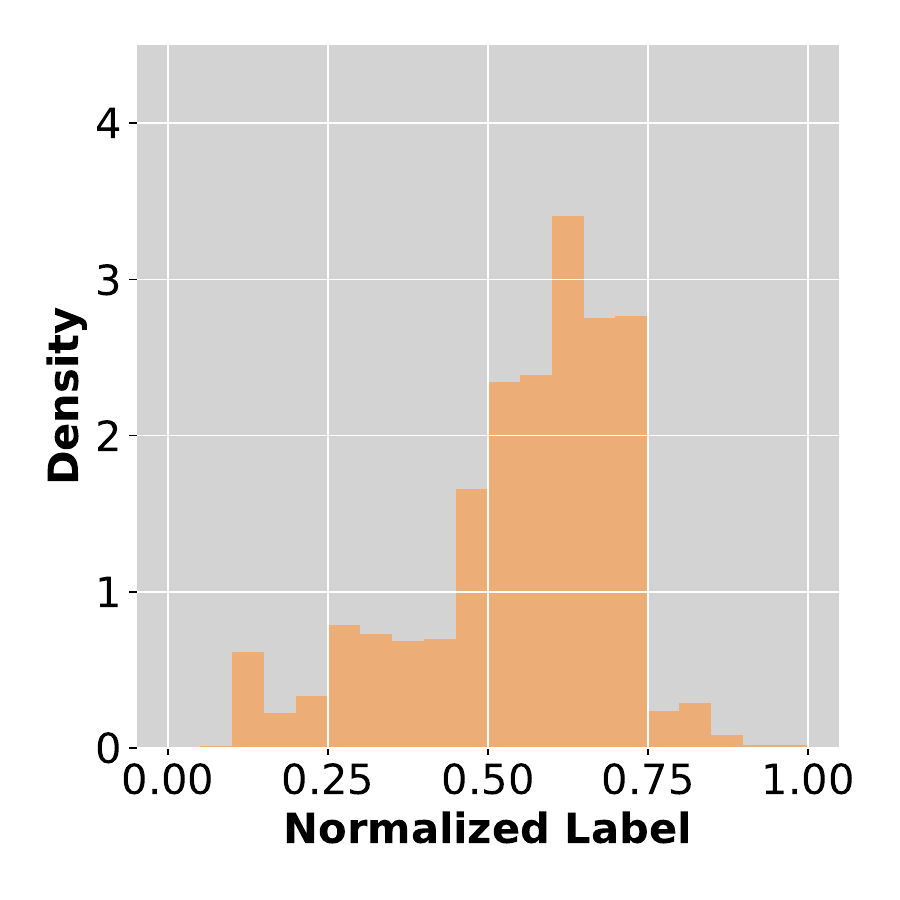}
    \caption{\textsf{\scriptsize{ULTRA8T}}}
    \label{fig:1e}
\end{subfigure}
\hfill
\begin{subfigure}[b]{0.16\textwidth}
    \setlength{\abovecaptionskip}{-2pt}
    \includegraphics[width=\textwidth]{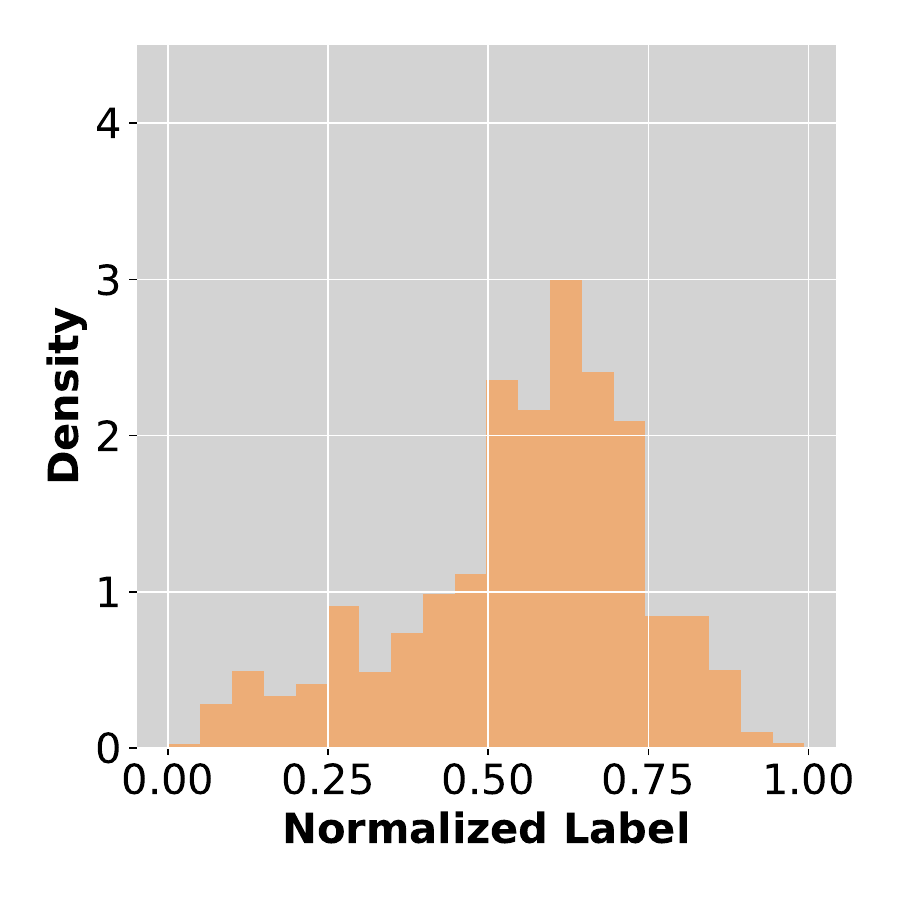}
    \caption{\textsf{\scriptsize{SANDWICH-RAM}}}
    \label{fig:1f}
\end{subfigure}
\caption{Normalized label distributions of all AMS circuit datasets.}
\label{fig:six_distr}
\end{figure*}

In parasitic estimation, label imbalance refers to the distributions of parasitic capacitance that are heavily skewed with long-tailed patterns in AMS circuits, as shown in \autoref{fig:six_distr}. However, our trained GNNs are used to other unseen designs. 
Such an imbalance degrades the generalizability and fairness of GNNs, and it poses a critical challenge that demands dedicated efforts to enhance data-driven transferability.
Balanced MSE \cite{ren2022balanced} and balanced softmax cross-entropy (BSCE) \cite{ren2020balanced} address the distribution mismatch between training and test circuits by serving as statistically principled loss functions. Assuming the test set labels follow a balanced distribution with conditional probability $p_{\text{bal}}(\boldsymbol{y} | \boldsymbol{x})$, we derive $p_{\text{bal}}(\boldsymbol{y} | \boldsymbol{x})$ from the skewed training distribution $p_{\text{train}}(\boldsymbol{y} | \boldsymbol{x})$ using the training label distribution $p_{\text{train}}(\boldsymbol{y})$.  
Expanding \autoref{eq:change_variable}, the relationship between the training and balanced distributions is expressed as:  
\begin{equation}\label{eq:inst_reg}  
p_{\text{train}}(\boldsymbol{y} | \boldsymbol{x}) = \frac{p_{\text{bal}}(\boldsymbol{y} | \boldsymbol{x}) \cdot p_{\text{train}}(\boldsymbol{y})}{\int_{\mathbf{Y}} p_{\text{bal}}(\boldsymbol{y}' | \boldsymbol{x}) \cdot p_{\text{train}}(\boldsymbol{y}') \, d\boldsymbol{y}'}.  
\end{equation}  
To estimate $p_{\text{bal}}(\boldsymbol{y} | \boldsymbol{x})$, we minimize the negative log-likelihood (NLL) of $p_{\text{train}}(\boldsymbol{y} | \boldsymbol{x})$ (see \cite{ren2022balanced} for proof). During training, we:  
(i) compute $p_{\text{bal}}(\boldsymbol{y} | \boldsymbol{x}; \boldsymbol{\theta})$ using the regressor,  
(ii) convert it to $p_{\text{train}}(\boldsymbol{y} | \boldsymbol{x}; \boldsymbol{\theta})$ via \autoref{eq:inst_reg},  
(iii) update parameters $\boldsymbol{\theta}$ by minimizing the NLL loss.  
During inference, the regressor directly estimates $p_{\text{bal}}(\boldsymbol{y} | \boldsymbol{x})$ without conversion:  
\begin{align}\label{eq:p_bal}  
p_{\text{bal}}(\boldsymbol{y} | \boldsymbol{x}; \boldsymbol{\theta}) = \mathcal{N}\left(\boldsymbol{y}; \boldsymbol{y}_{\text{pred}}, \sigma_{\text{noise}}^2 \mathbf{I}\right),  
\end{align}  
where $\boldsymbol{y}_{\text{pred}}$ is the model’s prediction.  

The balanced MSE and softmax CE losses are both derived from \autoref{eq:inst_reg} and \autoref{eq:p_bal}, with the former amending the MSE loss and the latter scaling logits to reflect test-time class balance. They share a common theoretical foundation in distribution alignment but differ in their handling of continuous vs. discrete labels.  

\subsection{Balanced MSE for Regression}\label{sec:bmse}

\begin{definition}[Balanced MSE]  
Given a regressor's prediction $\boldsymbol{y}_{\text{pred}}$ and the training label distribution prior $p_{\text{train}}(\boldsymbol{y})$, the balanced MSE (BMSE) loss is defined as:
\begin{equation}
\begin{split}\label{eq:balanced_mse}
\mathcal{L}_{\text{BMSE}} &= -\log p_{\text{train}}(\boldsymbol{y} | \boldsymbol{x}; \boldsymbol{\theta}) \\
&= -\log \frac{p_{\text{bal}}(\boldsymbol{y} | \boldsymbol{x}; \boldsymbol{\theta}) \cdot p_{\text{train}}(\boldsymbol{y})}{\int_{\mathbf{Y}} p_{\text{bal}}(\boldsymbol{y}' | \boldsymbol{x}; \boldsymbol{\theta}) \cdot p_{\text{train}}(\boldsymbol{y}') \, d\boldsymbol{y}'} \\
&\cong -\log \mathcal{N}(\boldsymbol{y}; \boldsymbol{y}_{\text{pred}}, \sigma_{\text{noise}}^2 \mathbf{I}) \\
&\quad + \log \int_{\mathbf{Y}} \mathcal{N}(\boldsymbol{y}'; \boldsymbol{y}_{\text{pred}}, \sigma_{\text{noise}}^2 \mathbf{I}) \cdot p_{\text{train}}(\boldsymbol{y}') \, d\boldsymbol{y}',
\end{split}
\end{equation}
where $\cong$ omits the constant term $-\log p_{\text{train}}(\boldsymbol{y})$.  
\end{definition}

The balanced MSE loss comprises two components:  
a standard MSE loss (first term), derived from the negative log-likelihood of the Gaussian prediction; a balancing term (second term), which corrects for label distribution skew by integrating over the entire label space $\mathbf{Y}$.  
As demonstrated by Ren et al. \cite{ren2022balanced}, the standard MSE loss emerges as a special case of balanced MSE when $p_{\text{train}}(\boldsymbol{y})$ is uniform.

To operationalize \autoref{eq:balanced_mse}, the integral term must be evaluated in closed form. A key challenge lies in modeling $p_{\text{train}}(\boldsymbol{y})$ to ensure tractability. We propose two implementations:  

\noindent{\textbf{GMM-based Analytical Integration (GAI)}}.  
Assume $p_{\text{train}}(\boldsymbol{y})$ follows a Gaussian Mixture Model (GMM), which enables analytical integration due to the closure property of Gaussians under multiplication. Let:  
\begin{align}\label{eq:gmm}  
p_{\text{train}}(\boldsymbol{y}) = \sum_{i=1}^K \phi_i \mathcal{N}(\boldsymbol{y}; \boldsymbol{\mu}_i, \boldsymbol{\Sigma}_i),  
\end{align}  
where $K$ is the number of components, and $\phi_i$, $\boldsymbol{\mu}_i$, $\boldsymbol{\Sigma}_i$ denote the weight, mean, and covariance of the $i$-th Gaussian. Substituting \autoref{eq:gmm} into \autoref{eq:balanced_mse} yields:  
\begin{equation}  
\begin{split}\label{eq:gmm_loss_multi}  
\mathcal{L}_{\text{GAI}} &= -\log \mathcal{N}(\boldsymbol{y}; \boldsymbol{y}_{\text{pred}}, \sigma_{\text{noise}}^2 \mathbf{I}) \\  
&\quad + \log \sum_{i=1}^K \phi_i \mathcal{N}(\boldsymbol{y}_{\text{pred}}; \boldsymbol{\mu}_i, \boldsymbol{\Sigma}_i + \sigma_{\text{noise}}^2 \mathbf{I}).  
\end{split}  
\end{equation}  
The second term approximates the integral via GMM components, leveraging the conjugacy of Gaussians.  

\noindent{\textbf{Batch-based Monte Carlo (BMC)}}.
With batch size $N$, BMC estimates the integral empirically using labels in a training batch $\mathcal{B}_{\boldsymbol{y}} = \{\boldsymbol{y}^{(1)}, \boldsymbol{y}^{(2)}, \ldots, \boldsymbol{y}^{(N)}\}$, requiring no prior knowledge of $p_{\text{train}}(\boldsymbol{y})$. The loss becomes:  
\begin{equation}  
\begin{split}\label{eq:mcm_batch}  
\mathcal{L}_{\text{BMC}} &= -\log \mathcal{N}(\boldsymbol{y}; \boldsymbol{y}_{\text{pred}}, \sigma_{\text{noise}}^2 \mathbf{I}) \\  
&\quad + \log \sum_{i=1}^N \mathcal{N}(\boldsymbol{y}^{(i)}; \boldsymbol{y}_{\text{pred}}, \sigma_{\text{noise}}^2 \mathbf{I}).  
\end{split}  
\end{equation}  
By rewriting \autoref{eq:mcm_batch}, we can see its connection to temperature-scaled softmax:
\begin{align}\label{eq:bmc_loss}  
\mathcal{L} = -\log \frac{\exp\left(-\|\boldsymbol{y}_{\text{pred}} - \boldsymbol{y}\|_2^2 / \tau\right)}{\sum_{\boldsymbol{y}' \in \mathcal{B}_{\boldsymbol{y}}} \exp\left(-\|\boldsymbol{y}_{\text{pred}} - \boldsymbol{y}'\|_2^2 / \tau\right)},  
\end{align}  
where $\tau = 2\sigma_{\text{noise}}^2$ controls the sharpness of the weighting. 

\subsection{Balanced Softmax for Classification}  
In imbalanced classification, where the label space $\mathcal{Y}$ is discrete and one-dimensional, the relationship between training and balanced distributions (\autoref{eq:inst_reg}) remains valid but replaces integration with summation:  
\begin{align}\label{eq:inst_cls}  
p_{\text{train}}(y | \boldsymbol{x}) = \frac{p_{\text{bal}}(y | \boldsymbol{x}) \cdot p_{\text{train}}(y)}{\sum_{y' \in \mathcal{Y}} p_{\text{bal}}(y' | \boldsymbol{x}) \cdot p_{\text{train}}(y')},  
\end{align}  
where $y \in \mathcal{Y}$ denotes discrete class labels, and $p_{\text{train}}(y)$ is the empirical frequency of class $y$ in the training set. 
Moreover, the balanced distribution $p_{\text{bal}}(y | \boldsymbol{x}; \boldsymbol{\theta})$ is typically modeled via softmax normalization for classification:  
\begin{align}\label{eq:softmax}  
p_{\text{bal}}(y | \boldsymbol{x}; \boldsymbol{\theta}) = \frac{\exp(\eta[y])}{\sum_{y' \in \mathcal{Y}} \exp(\eta[y'])},  
\end{align}  
where $\eta[y] \in \mathbb{R}$ is the logit (unnormalized score) for class $y$. Substituting \autoref{eq:softmax} into \autoref{eq:inst_cls} yields:  
\begin{align}\label{eq:logit_adjustment}  
p_{\text{train}}(y | \boldsymbol{x}; \boldsymbol{\theta}) = \frac{\exp(\eta[y]) \cdot p_{\text{train}}(y)}{\sum_{y' \in \mathcal{Y}} \exp(\eta[y']) \cdot p_{\text{train}}(y')}.  
\end{align}  
This aligns with logit adjustment techniques in imbalanced classification \cite{ren2020balanced, menon2020long, hong2021disentangling}, which offset logits by class frequency logarithms. 

\begin{definition}[Balanced Softmax CE]  
Given a classifier's raw logits $\eta[y]$ and the training label distribution prior $p_{\text{train}}(y)$, the balanced Softmax cross-entropy loss is defined as:
\begin{equation}\label{eq:bce_loss} 
\begin{split}
\mathcal{L}_{\text{BSCE}} &= -\log \frac{\exp(\eta[y]) \cdot p_{\text{train}}(y)}{\sum_{y' \in \mathcal{Y}} \exp(\eta[y']) \cdot p_{\text{train}}(y')} \\
&= -(\eta[y] + \log p_{\text{train}}(y)) \\
&\quad + \log\left(\sum_{y' \in \mathcal{Y}} \exp(\eta[y'] + \log p_{\text{train}}(y'))\right).
\end{split}
\end{equation}
\end{definition}

CircuitGCL provides a unified view of imbalanced regression and classification. By deriving BMSE and BSCE from the same distribution alignment principle, we validate these loss functions as dual instantiations of a core theoretical insight.




\begin{table}[!tb]
\centering
\caption{AMS Circuit Dataset Statistics. One Design Case Is Sufficient to Train CircuitGCL.}
\label{tab:dataset}
    \begin{tabular}{c|c|ccc}
    \toprule
    Split & Dataset & $N$ & $N_E$ & \#Links  \\ \midrule
    \multirow{1}{*}{Train.\&Val.} 
    & \ssram & 87K & 134K & 131K  \\
    \midrule
    \multirow{5}{*}{Test} 
    & \digtime & 17K & 36K & 4K \\
    & \timectrl & 18K & 44K & 5K  \\
    & \sarray & 144K & 352K & 110K \\ 
    & \ultra & 3.5M & 13.4M & 166K \\ 
    & \sandwich & 4.3M & 13.3M & 154K \\
    \bottomrule
    \end{tabular}
\end{table}

\section{Experiments}
Our implementation leverages PyG \cite{FeyLenssen2019PyG} for graph processing. All experiments were conducted on a shared computing cluster equipped with 40 Intel Xeon Silver 4314 CPUs (2.4 GHz), 128GB RAM, and four NVIDIA RTX 4090 GPUs (24GB VRAM). Each training run utilized 4-6 CPU cores, one GPU, and 128GB of memory. In data preparation, full schematic netlists were first parsed to extract graph structures and node features (circuit statistics), which is the same as work \cite{shen2025few-shot}; then, post-layout netlists (Standard Parasitic Format/SPF files) were also processed to collect ground-truth coupling capacitance labels. The subgraph sampling is implemented by ``LinkNeigbhorLoader" in PyG.
We compare CircuitGCL against three state-of-the-art approaches for parasitic capacitance prediction:  
(i) ParaGraph \cite{ParaGraph} is a MPNN-based ensemble model;  
(ii) DLPL-Cap \cite{shen2024deep} is a multi-expert GNN regressor; 
(iii) CircuitGPS \cite{shen2025few-shot} uses few-shot learning with positional encoding.

\subsection{AMS Datasets}

\begin{table*}[!tb]
\caption{Error Comparison of CircuitGCL and Prior Methods on Edge Regression.}
\label{tab:reg_comp}
\setlength\tabcolsep{2.8pt}
\resizebox{\linewidth}{!}{
    \begin{tabular}{l|cccc|cccc|cccc|cccc}
    \toprule
    Testset & \multicolumn{4}{c|}{\timectrl} & \multicolumn{4}{c|}{\sarray} & \multicolumn{4}{c|}{\ultra} & \multicolumn{4}{c}{\sandwich} \\
    \rowcolor{gray!40}
    Metric & Loss & MAE$\downarrow$ & MSE$\downarrow$ & $R^2\uparrow$ & Loss & MAE$\downarrow$ & MSE$\downarrow$ & $R^2\uparrow$ & Loss & MAE$\downarrow$ & MSE$\downarrow$ & $R^2\uparrow$ & Loss & MAE$\downarrow$ & MSE$\downarrow$ & $R^2\uparrow$ \\  
    \midrule
    ParaGraph & 0.0153 & 0.0914 & 0.0153 & 0.5250 & 0.0115 & 0.0788 & 0.0115 & 0.4252 & 0.0175 & 0.0937 & 0.0175 & 0.3200 & 0.0223 & 0.1087 & 0.0223 & 0.3389 \\
    CircuitGPS & 0.0105 & 0.0742 & 0.0105 & 0.6911 & 0.0108 & 0.0701 & 0.0108 & 0.4576 & 0.0158 & 0.0818 & 0.0158 & 0.3845 & 0.0225 & 0.1039 & 0.0225 & 0.3326 \\
    DLPL-Cap & 0.0093 & 0.0701 & 0.0093 & 0.7056 & 0.0123 & 0.0806 & 0.0123 & 0.3853 & 0.0160 & 0.0813 & 0.0160 & 0.3704 & 0.0214 & 0.1012 & 0.0214 & 0.3622 \\ \midrule
    \begin{tabular}[c]{@{}l@{}}CircuitGCL \\ (MSE)\end{tabular} & 0.0118 & 0.0868 & 0.0118 & 0.6521 & 0.0093 & 0.0671 & 0.0093 & 0.5350 & 0.0144 & 0.0794 & \textbf{0.0144} & \textbf{0.4398} & 0.0193 & 0.0992 & 0.0193 & 0.4280 \\
    \begin{tabular}[c]{@{}l@{}}CircuitGCL \\ (BMC)\end{tabular} &  71.626 & 0.0628 & \textbf{0.0088} & \textbf{0.7407} & 74.313 & \textbf{0.0650} & 0.0092 & 0.5418 & 117.73 & 0.0771 & 0.0145 & 0.4350 & 152.34 & 0.0938 & 0.0188 & 0.4422 \\
    \begin{tabular}[c]{@{}l@{}}CircuitGCL \\ (GAI)\end{tabular} &  0.0091 & \textbf{0.0610} & 0.0091 & 0.7300 & 0.0089 & 0.0667 & \textbf{0.0089} & \textbf{0.5556} & 0.0145 & \textbf{0.0762} & 0.0145 & 0.4358 & 0.0187 & \textbf{0.0935} & \textbf{0.0187} & \textbf{0.4445} \\
    \midrule
    Max. Impr. &  - & 33.26\% & 42.48\% & 41.08\% & - & 19.35\% & 27.64\% & 44.20\% & - & 18.68\% & 17.71\% & 37.44\% & - & 0.1398\% & 16.89\% & 33.64\% \\
    \bottomrule
    \end{tabular}
}
\end{table*}

\begin{table*}[tb]
\caption{Accuracy Comparison of CircuitGCL and Prior Methods on Node Classification.}
\label{tab:acc_comp}
\setlength\tabcolsep{1pt}
\resizebox{\linewidth}{!}{
    \begin{tabular}{l|ccccc|ccccc|ccccc|ccccc}
    \toprule
    Test Set & \multicolumn{5}{c|}{\digtime} & \multicolumn{5}{c|}{\sarray} & \multicolumn{5}{c|}{\ultra} & \multicolumn{5}{c}{\sandwich} \\
    \rowcolor{gray!40}
    Metric & Loss & Acc.$\uparrow$ & Precision$\uparrow$ & Recall$\uparrow$ & F1$\uparrow$ & Loss & Acc.$\uparrow$ & Precision$\uparrow$ & Recall$\uparrow$ & F1$\uparrow$ & Loss & Acc.$\uparrow$ & Precision$\uparrow$ & Recall$\uparrow$ & F1$\uparrow$ & Loss & Acc.$\uparrow$ & Precision$\uparrow$ & Recall$\uparrow$ & F1$\uparrow$ \\  
    \midrule
    ParaGraph & 0.1973 & 0.2359 & 0.1249 & 0.3128 & 0.1771 & 0.0730 & 0.6024 & 0.2923 & 0.3239 & 0.2945 & 0.3228 & 0.6252 & 0.3058 & 0.3129 & 0.2973 & 0.8511 & 0.3651 & 0.2485 & 0.2675 & 0.2199\\
    CircuitGPS & 3.1762 & 0.2082 & 0.2449 & 0.3119 & 0.2259 & 1.5075 & 0.2200 & 0.4000 & 0.1628 & 0.2299 & 1.0922 & 0.3925 & 0.4541 & 0.2506 & 0.2892 & 2.307 & 0.4185 & 0.4494 & 0.3573 & 0.3002\\
    DLPL-Cap & 2.0780   & \textbf{0.7130} & \textbf{0.6918} &  \textbf{0.6755} &  \textbf{0.6167} & 0.0142 &  0.8888 & 0.6943 & 0.6481 & 0.6578 & 0.3289 & 0.8683 &  0.6099 & 0.6109 & 0.5984 &  2.6968 & 0.5701 & 0.4783 &  0.4140 & 0.4248   \\ 
    \midrule
    \begin{tabular}[c]{@{}l@{}}CircuitGCL \\ (CE)\end{tabular} & 20.936 & 0.5400  & 0.3611 & 0.4121 & 0.3328 & 0.5537 & 0.6792 & 0.5037 & 0.5304 & 0.5174 & 0.9860  & 0.5787 & 0.6194 & 0.4316 & 0.4592 & 2.1311 & 0.4441 & 0.4990  & 0.3044 & 0.3607 \\
    \begin{tabular}[c]{@{}l@{}}CircuitGCL \\ (Focal)\end{tabular} & 3.9955  & 0.5380 & 0.3503 & 0.4021 & 0.3145 & 0.0226 & 0.8994 & 0.5756 & 0.5402 & 0.5533 & 0.4021 & 0.8163 & \textbf{0.6381} & 0.6040  & 0.6157 & 1.2232 & 0.5809 & 0.5227 & 0.4210  & 0.4580  \\
    \begin{tabular}[c]{@{}l@{}}CircuitGCL \\ (BSCE)\end{tabular} & 18.898 &  0.6760 & 0.6855 & 0.5981 & 0.5309 &  0.4989 &  \textbf{0.9622} &  \textbf{0.7195} & \textbf{0.7211} & \textbf{0.7185} & 0.9616 &  \textbf{0.9146} & 0.6376 &  \textbf{0.6391} &  \textbf{0.6382} & 1.7694 & \textbf{0.7231} &  \textbf{0.5826} &  \textbf{0.5392} &  \textbf{0.5514} \\
    \midrule
    Max. Impr. & - & 2.2$\times$ & 4.5$\times$ & 0.9$\times$ & 0.9$\times$ & - & 3.4$\times$ & 0.8$\times$ & 3.4$\times$ & 2.1$\times$ & - & 1.3$\times$ & 1.1$\times$ & 1.6$\times$ & 1.2$\times$ & - & 1.0$\times$ & 1.3$\times$ & 1.0$\times$ & 1.5$\times$ \\
    \bottomrule
    \end{tabular}
}
\end{table*}

Table \ref{tab:dataset} summarizes the AMS circuit datasets used in our experiments, all implemented in TSMC 28nm CMOS technology. To demonstrate data transferability, we train CircuitGCL on a single mid-sized design and evaluate it on four unseen test cases. The graph contrastive learning (GCL) component is also pretrained on the same single design to learn expressive node representations.

\textsf{\footnotesize{SSRAM}} \cite{tscache} is a medium-scale energy-efficient design combining standard digital cells and SRAM arrays.  
\textsf{\footnotesize{SANDWICH-RAM}} \cite{cim1} features a balanced architecture with computational digital circuits and storage SRAM arrays in alternating layers.  
\textsf{\footnotesize{ULTRA8T}} SRAM \cite{shen2024ultra8t} is the largest design with multi-voltage domains and extensive analog modules.  
As for test sets, \textsf{\footnotesize{DIGITAL\_CLK\_GEN}} is a digital/SRAM hybrid for internal SRAM clock generation.  
\textsf{\footnotesize{TIMING\_CONTROL}} is a digital control signal generator for SRAM operations.  
\textsf{\footnotesize{ARRAY\_128\_32}} is a standalone 128×32 SRAM array.  
All test sets are strictly excluded from training/validation data, ensuring zero-shot evaluation. 

\autoref{fig:six_distr} illustrates the significant distribution shifts between datasets, highlighting the challenge of developing universal transfer learning methods for such AMS circuits.  In the regression task, we keep the coupling capacitor with capacitance $y_i \in [1^{-21}, 1^{-15}]$. In the classification task, we divide the ground capacitors connecting to net nodes into 5 categories in terms of the magnitude of their capacitance values.

\subsection{Edge Regression - Coupling Capacitance Estimation}
\label{sec:com_reg}
In this task, we configure encoders in CircuitGCL with 4-layer ClusterGCN \cite{chiang2019cluster}, 256 hidden dimensions, Tanh activation, 0.3 dropout, and $1^{-6}$ learning rate. The downstream GNN adopts a 5-layer GraphSAGE \cite{hamilton2017inductive}, with 144 hidden dimensions, PReLU activation, 0.3 dropout, and $1^{-4}$ learning rate. The task-specific heads are 2-layer MLPs with the same number of hidden dimensions as the CL encoders and the downstream GNN, respectively. The $\sigma_{noise}$ in \autoref{eq:gmm_loss_multi} and \autoref{eq:bmc_loss} is set to be 0.001.

Table \ref{tab:reg_comp} compares CircuitGCL against prior methods on parasitic capacitance regression. GAI variant achieves the best overall performance across all test sets. On large-scale datasets, GAI significantly outperforms existing methods—for example, 23\% improvement over DLPL-Cap on \sandwich. The technique shows consistent advantages on both small (\timectrl) and large (\ultra, \sandwich) designs, while prior approaches struggle with scalability. Although BMC exhibits high absolute loss values due to batch normalization, it maintains competitive relative metrics. These results validate CircuitGCL's transferable representation learning capability for parasitic prediction.

\begin{figure*}[!t]
    \centering
    \begin{minipage}{0.49\textwidth}
    \includegraphics[width=0.86\linewidth]{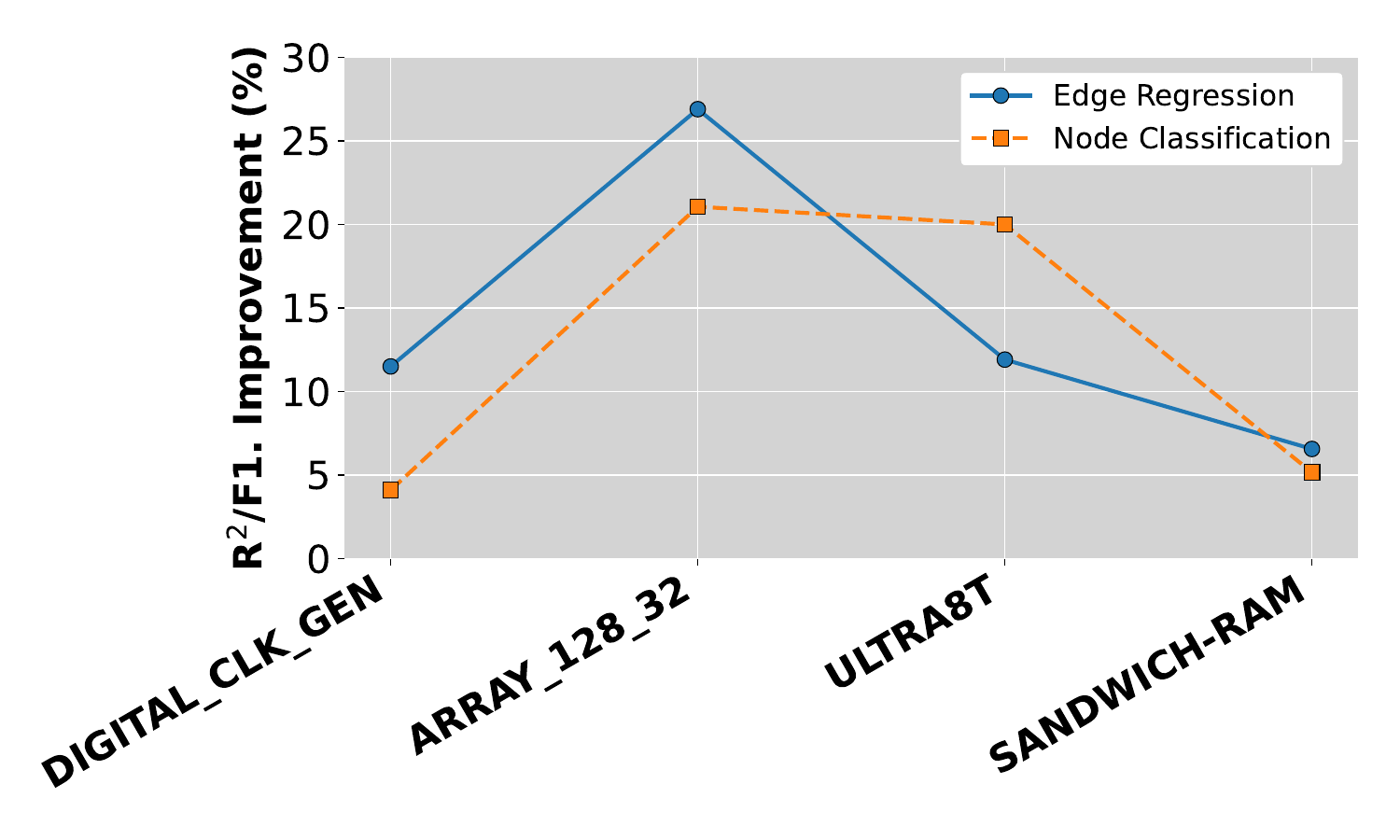}
    \caption{$R^2$ and F1 improvements of applying RSM to regression and classification tasks, respectively.}
    \label{fig:ext_rsm}
    \end{minipage}
    \hfill
    \begin{minipage}{0.49\textwidth}
    \includegraphics[width=0.86\linewidth]{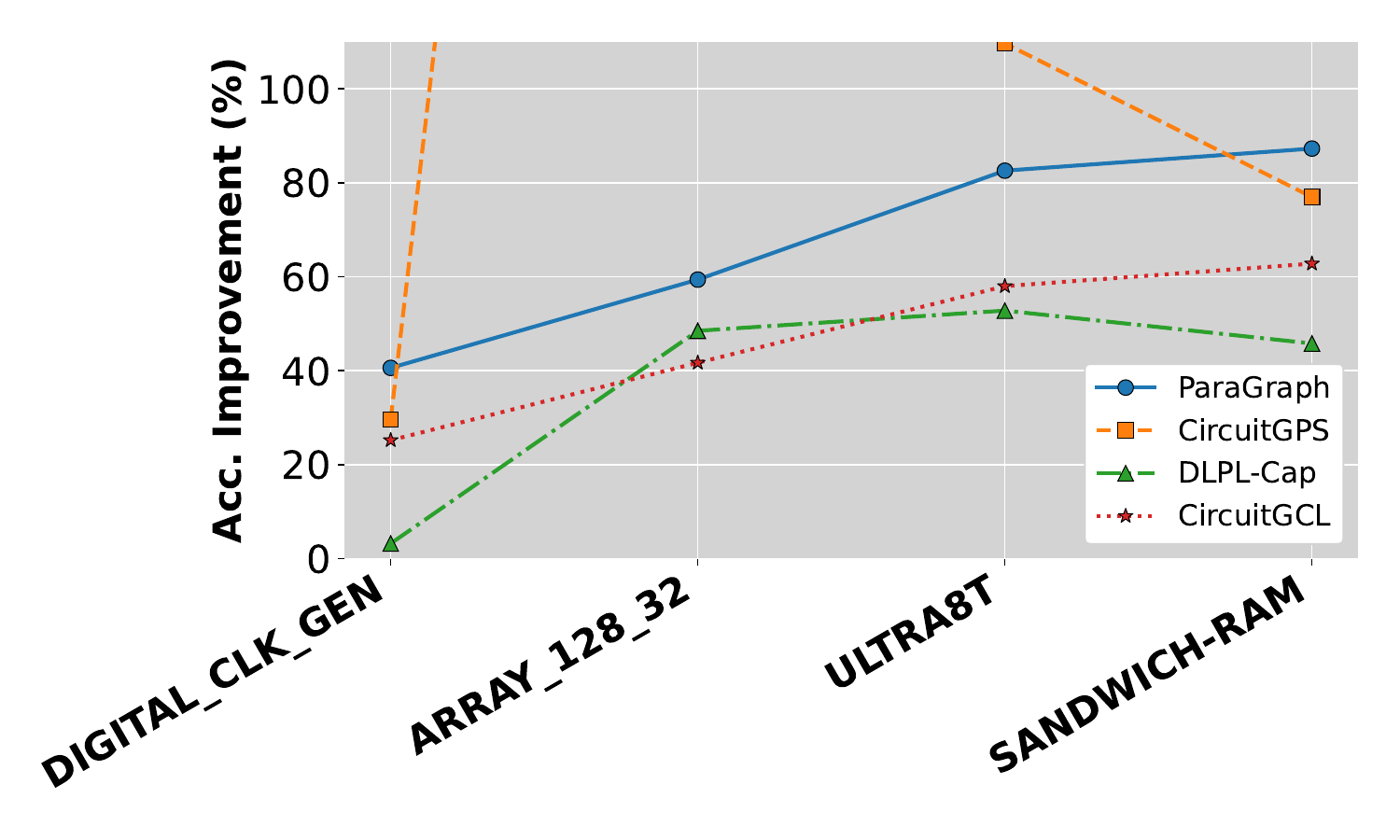}
    \caption{Accuracy improvement of applying BSCE to other baselines. The accuracy gains are normalized to CE.}
    \label{fig:ext_bsm}
    \end{minipage}
\end{figure*}

\begin{figure}[tbp]
\vspace{-4pt}
\setlength{\belowcaptionskip}{-6pt}
    \centering
    \includegraphics[width=\linewidth]{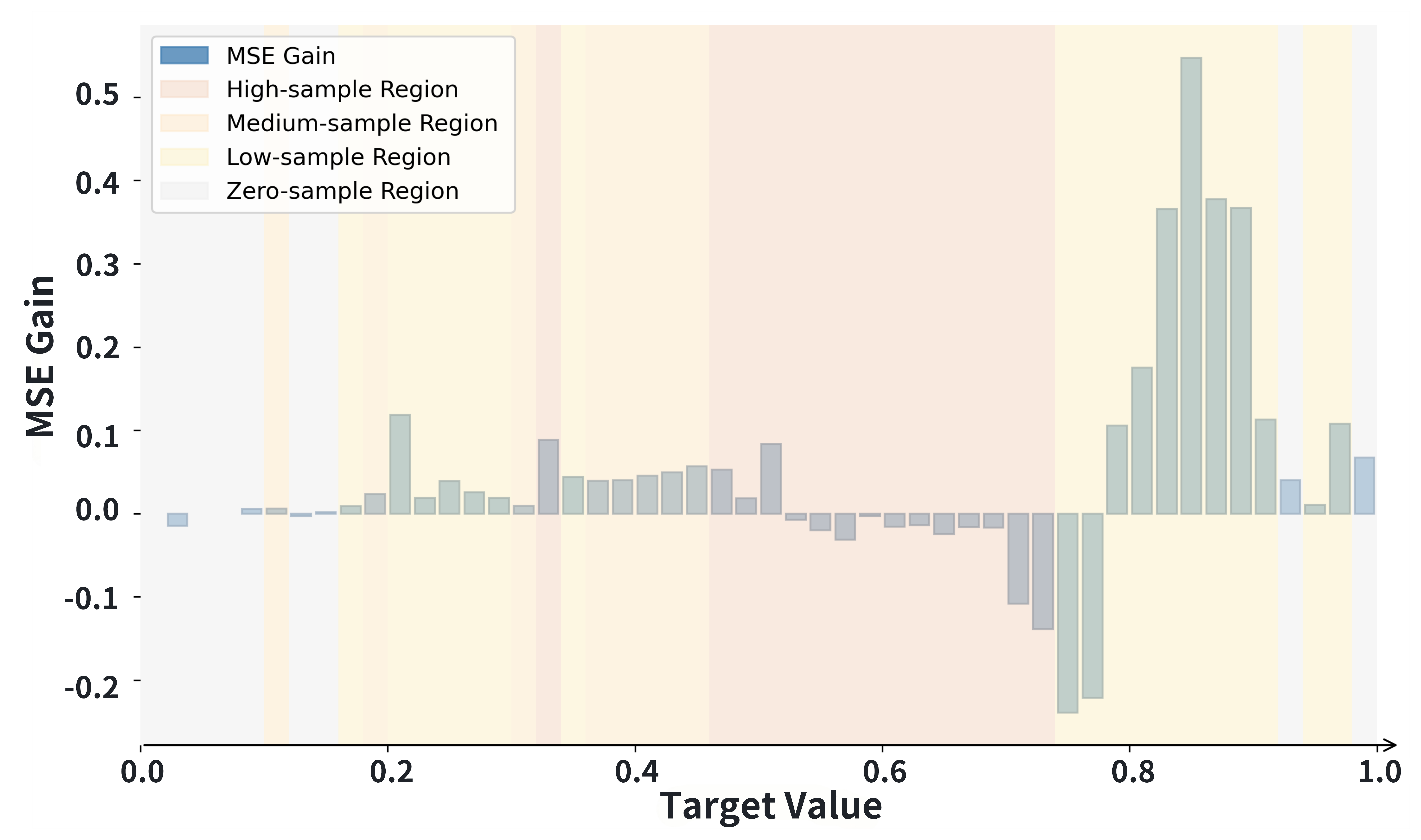}
    
    \caption{MSE improvement through label rebalancing technique. Balanced MSE significantly enhances model performance in data-scarce regions (pale yellow background).}
    \label{fig:laebel_rebalancing_constrast}
\end{figure}

\subsection{Node Classification - Ground Capacitance Classification}

In node classification, we configure encoders as in edge regression (\autoref{sec:com_reg}) and use 4-layer GraphSAGE with focal loss for fair comparison. We categorize net nodes into 5 classes based on normalized ground capacitance (\autoref{fig:six_distr}), excluding the most frequent category (with label 2) when calculating metrics to emphasize model differences.

Table \ref{tab:acc_comp} compares CircuitGCL against baselines on node classification. CircuitGCL with BSCE achieves state-of-the-art performance, particularly on large-scale designs with significant improvements over DLPL-Cap (e.g., 3.4× accuracy improvement on \sarray). While DLPL-Cap excels on \digtime, it fails to generalize to larger designs, highlighting its reliance on task-specific tuning. ParaGraph and CircuitGPS show poor scalability with limited F1 scores on \sandwich. CircuitGCL's balanced softmax CE effectively addresses label imbalance, achieving robust performance across diverse circuit topologies and demonstrating superior scalability over existing baselines.

\subsection{Extended Study on RSM of GCL}

Here, we further discuss the impact of using the RSM of contrastive learning.
As depicted in \autoref{fig:ext_rsm}, RSM provides the most significant $R^2$ improvement on \sarray~(26.9\%), suggesting it effectively handles the complex coupling patterns in SRAM arrays.
The largest F1 gain occurs on \ultra~(20.0\%), demonstrating RSM's ability to improve semantic clustering in large-scale designs.
While improvements vary by dataset, RSM never degrades performance, with minimum gains of 4.1\% (F1) and 6.56\% ($R^2$).
These results validate RSM as a crucial component of CircuitGCL for learning transferable representations in AMS circuits.


\subsection{Extended Study on Label Rebalancing}

To further demonstrate the effectiveness of balanced loss functions, we conduct an extended study using BSCE as the loss function for all baselines. As shown in \autoref{fig:ext_bsm}, BSCE yields significant accuracy improvements across all methods, with particularly substantial gains observed in large-scale designs, highlighting BSCE's effectiveness for scalable imbalanced classification. \autoref{fig:laebel_rebalancing_constrast} demonstrates the performance improvements of label rebalancing across data-scarce regions, where MSE gain shows it significantly enhances model performance in these challenging regions.


\section{Conclusion}
CircuitGCL's self-supervised paradigm and distribution-aware losses address two universal EDA challenges, particularly pronounced in AMS circuit design: data scarcity (via GCL) and label imbalance (via rebalancing). Its graph-native architecture aligns naturally with circuit netlists, enabling seamless adoption in commercial tools for tasks requiring rapid design-space exploration. 
This positions CircuitGCL as a foundational framework for next-generation EDA tools, bridging the gap between data-driven automation and precision-critical circuit design. Future work could extend the framework to broader AMS circuit varieties, adapt CircuitGCL to parasitic resistance estimation, or integrate it into an RC-aware placement \& routing tool.

\clearpage
\bibliographystyle{IEEEtran}
\bibliography{refs}

    
    

\end{document}